\documentclass[letterpaper, 10 pt, conference]{ieeeconf}  % Comment this line out if you need a4paper

\IEEEoverridecommandlockouts                              % This command is only needed if 
\usepackage{epsfig} % for postscript graphics files

\title{\LARGE \bf
FedBChain: A Blockchain-enabled Federated Learning Framework for Improving DeepConvLSTM with Comparative Strategy Insights*
}

\author{Gaoxuan Li$^{1}$, Chern Hong Lim$^{2}$, Qiyao Ma$^{3}$, Xinyu Tang$^{4}$, Hwa Hui Tew$^{5}$, Fan Ding$^{6}$, and Xuewen Luo$^{7}$% <-this % stops a space
\thanks{*This work was not supported by any organization}% <-this % stops a space
\thanks{$^{1}$Gaoxuan Li is with the School of Information Technology, Monash University, Selangor, Malaysia
        {\tt\small glii0053@student.monash.edu}}%
\thanks{$^{2}$Chern Hong Lim is with the School of Information Technology, Monash University, Selangor, Malaysia
        {\tt\small lim.chernhong@monash.edu}}%
\thanks{$^{3}$Qiyao Ma is with the Arts College, Sichuan University, Sichuan, China
        {\tt\small 1115744893@qq.com}}%
\thanks{$^{4}$Xinyu Tang is with the School of Information Technology, Monash University, Selangor, Malaysia
        {\tt\small xtan0132@student.monash.edu}}%
\thanks{$^{5}$Hwa Hui Tew is with the School of Information Technology, Monash University, Selangor, Malaysia
        {\tt\small hwa.tew@monash.edu}}%
\thanks{$^{6}$Fan Ding is with the School of Information Technology, Monash University, Selangor, Malaysia
        {\tt\small fdin0009@student.monash.edu}}%
\thanks{$^{7}$Xuewen Luo is with the School of Information Technology, Monash University, Selangor, Malaysia
        {\tt\small xluo0033@student.monash.edu}}%
}
\usepackage{tabularray}

\begin{document}

\maketitle
\thispagestyle{empty}
\pagestyle{empty}

%%%%%%%%%%%%%%%%%%%%%%%%%%%%%%%%%%%%%%%%%%%%%%%%%%%%%%%%%%%%%%%%%%%%%%%%%%%%%%%%
\begin{abstract}

Recent research in the field of Human Activity Recognition has shown that an improvement in prediction performance can be achieved by reducing the number of LSTM layers. However, this kind of enhancement is only significant on monolithic architectures, and when it runs on large-scale distributed training, data security and privacy issues will be reconsidered, and its prediction performance is unknown. In this paper, we introduce a novel framework: FedBChain, which integrates the federated learning paradigm based on a modified DeepConvLSTM architecture with a single LSTM layer. This framework performs comparative tests of prediction performance on three different real-world datasets based on three different hidden layer units (128, 256, and 512) combined with five different federated learning strategies, respectively. The results show that our architecture has significant improvements in Precision, Recall and F1-score compared to the centralized training approach on all datasets with all hidden layer units for all strategies: FedAvg strategy improves on average by 4.54\%, FedProx improves on average by 4.57\%, FedTrimmedAvg improves on average by 4.35\%, Krum improves by 4.18\% on average, and FedAvgM improves by 4.46\% on average. Based on our results, it can be seen that FedBChain not only improves in performance, but also guarantees the security and privacy of user data compared to centralized training methods during the training process. The code for our experiments is publicly available (https://github.com/Glen909/FedBChain).

\end{abstract}
%%%%%%%%%%%%%%%%%%%%%%%%%%%%%%%%%%%%%%%%%%%%%%%%%%%%%%%%%%%%%%%%%%%%%%%%%%%%%%%%
\section{INTRODUCTION}

Human activity recognition(HAR) [1] has a wide range of applications and importance in several fields. In the field of digital health, recognizing human activities, especially for the elderly, can help maintain or improve their health; in the field of smart home, HAR can help automation systems better coordinate collaboration among devices in the home; in the field of human-computer interaction, clearly recognizing human activities can better understand human intentions and make timely and accurate responses. In a world where machine learning and deep learning are becoming increasingly fast and iterative, a larger number of researchers are applying them to the field of HAR to empower the recognition function and make it more efficient.

Among these researchers, DeepConvLSTM proposed by [2] is one of the deep learning architectures that has attracted a lot of attention in the study of HAR. This innovative architecture incorporates convolutional and recurrent layers. It shows excellent performance and outstanding results on several real-world datasets (Opportunity [3] and Skoda Mini Checkpoint datasets [4]). Specifically, [2] designed the LSTM layers in the DeepConvLSTM architecture as a two-layer structure and 128 more hidden units in each layer. As pointed out by [5] and further confirmed by [6]. It is generally accepted that at least two LSTM layers are required for optimal modelling during sequential data processing in the field of HAR.

However, this was doubted and reassessed by [7], whose study used a single LSTM layer in a recurrent layer structure for model training. The results showed that the DeepConvLSTM structure based on a two-layer LSTM was outperformed by the single-layer DeepConvLSTM (with the LSTM module of the second layer removed) in terms of accuracy, recall, and F1-scores for four of the five datasets. Specifically, [7] achieved an average of 62\% using fewer trainable parameters across the architecture with this adaptation and reduced the overall training time by 48\%. Although [7] achieved remarkable success on the DeepConvLSTM architecture based on single-layer LSTM, it is only applicable to monolithic servers, and in the large-scale distributed training environments, even if the performance of monolithic architectures is optimized again, it can not effectively stifle the security and privacy issues of personal data. Therefore, in order to better handle the privacy and security of user data during training process, federated learning [8] is proposed as an effective solution in large-scale distributed training environments. Federated learning is a decentralized machine learning approach that enables collaborative model training across multiple data owners, ensuring that data remains stored locally with each owner.

Therefore, this project develops FedBChain, a federated learning framework specialized for large-scale distributed training environments, on top of DeepConvLSTM. This framework employs distributed training approach instead of the centralized training approach, thus enhancing data privacy and security. In addition, compared to the general federated learning process, FedBChain employs blockchain technology [9] as the global model in designing the global model. Generally, federated learning usually relies on a central node for aggregation and distribution of model parameters, which may have a single point of failure that risks the overall system performance, therefore, the design of this framework uses blockchain technology instead of a centralized global model. Meanwhile, the deployment and execution of smart contracts automatically performs model aggregation and distribution of the latest parameters to further ensure the robustness and availability of the system.

The contributions of this paper are summarized as follows:
\begin{itemize}

\item This project introduces the FedBChain framework, which utilizes a single-layer DeepConvLSTM architecture for federated learning. This approach enhances data security and privacy throughout the training process. The framework is well-suited for large-scale distributed federated training.
\item Within the FedBChain framework, this project integrates a blockchain module to replace the centralized global model. This modification minimizes the risk of single point of failure and ensures that each model parameter update is securely recorded on the blockchain, making it traceable, reproducible and tamper-proof.
\item Based on the proposed FedBChain framework, this project analyses the DeepConvLSTM under five different federated learning strategies in comparison with centralized training methods, and tests the DeepConvLSTM on three datasets with three different configurations of hidden layer units. The results show that FedBChain significantly outperforms the centralized training method.

\end{itemize}
\section{RELATED WORK}

\subsection{Federated Learning}

Federated Learning (FL) has risen as a crucial framework for decentralized machine learning, designed to address data privacy issues while leveraging data from diverse sources for model training. Initially introduced by [10] in 2016, this approach ensures that data remains on local devices, with only model updates being shared to a central server, thereby significantly diminishing the risks associated with centralized data storage. Subsequent research has broadened to tackle various challenges in FL, including minimizing communication overhead as discussed by [11], ensuring equitable resource distribution among participants as explored by [12], and fortifying model security against adversarial attacks in distributed environments as identified by [13]. Additionally, advancements have been made with techniques that employ sophisticated encryption methods like secure multi-party computation and differential privacy to enhance the privacy protections of federated models, as demonstrated by [14].

\subsection{Human Activity Recognition}

HAR has garnered significant attention due to its applications in health monitoring and automated systems. Traditionally, HAR systems have relied heavily on sensor data, which is aggregated and processed on centralized servers [15]. However, privacy concerns and the logistical complexities associated with centralized data aggregation have spurred the exploration of decentralized methodologies, particularly through federated learning. For example, [16] illustrated the application of federated learning in HAR using mobile sensors, substantially enhancing privacy by keeping sensitive data on the devices themselves. More recent studies have furthered the field by integrating deep learning techniques to boost recognition accuracy while maintaining the decentralized framework of federated learning [17]. These developments suggest a shift toward more secure and efficient methods for implementing HAR in privacy-sensitive environments.

\subsection{Blockchain}

In the realm of HAR, the integration of blockchain technology has surfaced as an innovative trend aimed at bolstering system security and privacy. For example, [18] showcased a blockchain-based HAR system that employs smart contracts to verify and log sensor data, thereby enhancing transparency and trust in data management. [19] investigated how blockchain could safeguard user privacy through its distributed ledger technology, mitigating risks linked to centralized data storage. Meanwhile, [20] merged machine learning with blockchain to strengthen the immutability and automation of verification processes for HAR data on smart devices. Collectively, these studies not only underscore blockchain's potential to fortify privacy and security but also illustrate its role in improving the efficiency and transparency of data management.

\section{METHODS}

\subsection{Datasets} To evaluate the effectiveness of the DeepConvLSTM architecture after integrating the federated learning strategy, this project uses three well-known datasets processed in the same way as [7]: the Heterogeneous Activity Recognition (HHAR) dataset [21], the Real-World HAR (RWHAR)[22], and the Smartphone-based Human Activity and Pose Transition Recognition ( SBHAR) [23].

RWHAR: The RWHAR dataset contains data from 15 participants performing eight different activities (going up stairs, going down stairs, jumping, lying down, standing, sitting, running, and walking), as well as an empty class. According to the experimental parameters mentioned in [7], the present analysis was limited to 3D acceleration data recorded at 50Hz using wrist-worn sensors.

SBHAR: The SBHAR dataset contains data from 30 individuals engaged in daily activities such as standing, sitting, lying down, walking, walking up stairs and walking down stairs as shown in [23]. In addition to these six activities, the dataset includes six types of postural transitions (from standing to sitting, from sitting to standing, from sitting to lying, from lying to sitting, from standing to lying, and from lying to standing) and an empty category. To maintain comparability, this analysis also focuses only on the raw 3D acceleration data collected at a 50Hz sampling rate.

HHAR: Similar to the RWHAR dataset, the HHAR dataset contains data from nine individuals involved in daily activities, including six activities (cycling, sitting, standing, walking, and climbing up and down stairs) and the null category for prediction purposes. Consistent with the methodology of the previous dataset, the current analysis focused solely on 3D acceleration data recorded at 100Hz by a wrist-worn sensor unit.

   \begin{figure*}[thpb]
      \centering
      \includegraphics[width=\textwidth]{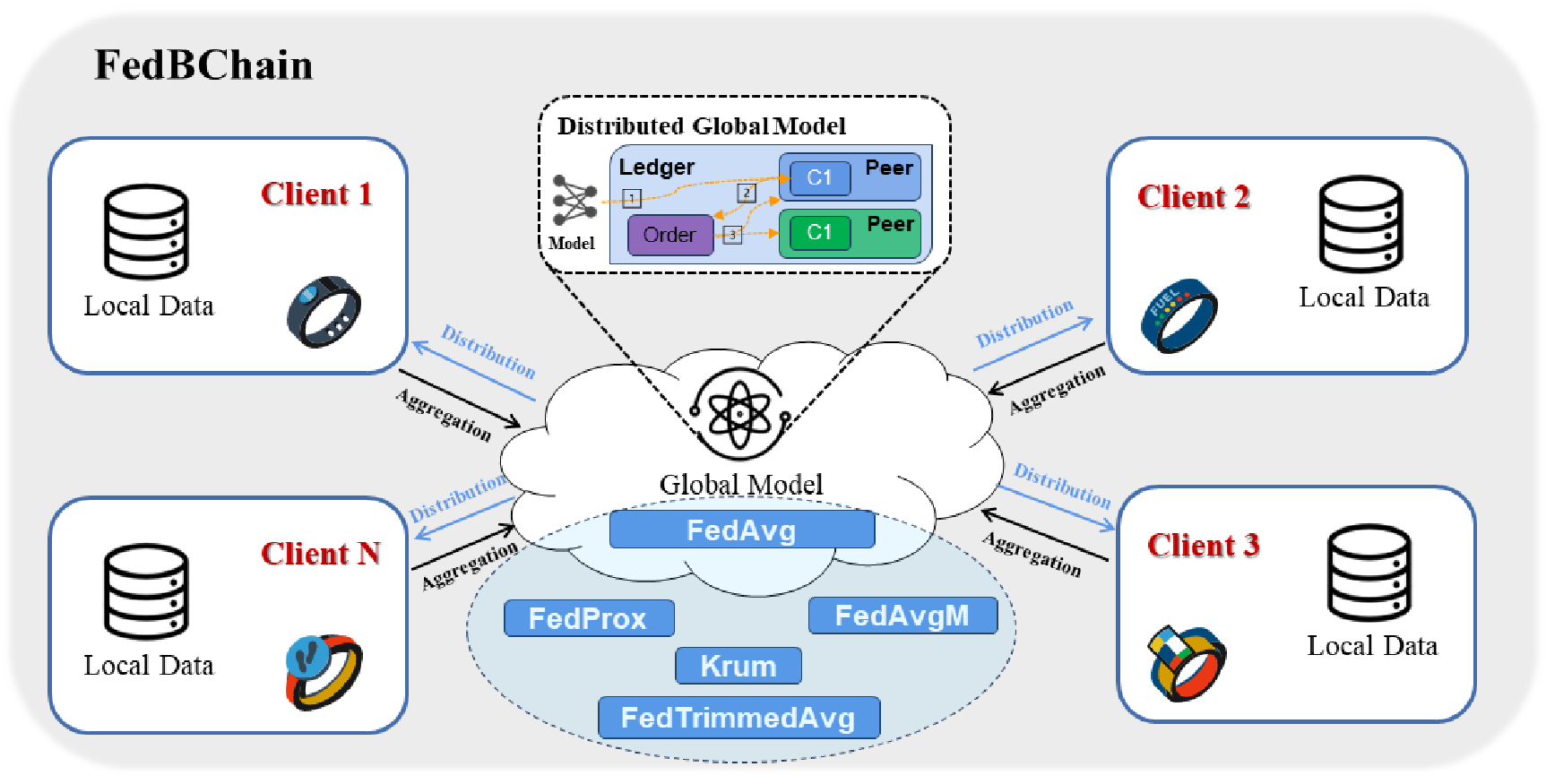} % scale adjust width width=\textwidth
      \caption{OVERVIEW OF FEDBCHAIN BASED ON FEDERATED LEARNING}
   \end{figure*}

\subsection{Architecture}
The novel framework, FedBChain, is illustrated in Fig. 1 and is structured around M servers and N clients. The M servers collaboratively establish a global model that is tasked with gathering model parameters, which are locally trained by the clients. Following specific strategies, these parameters are aggregated, and the resultant model parameters are subsequently redistributed to each client. The local models at the client sites then update their parameters from the global model before continuing training on local data. This cycle repeats until the model reaches convergence. Throughout this process, local data of the clients remains exclusively local and only weight parameters of the model are transmitted to the global model for aggregation.

Specifically, the FedBChain leverages the Flower framework [24] for implementation, where model parameter aggregation tasks are managed using FedAvg [25], FedProx [26], FedTrimmedAvg [27], Krum [28], and FedAvgM [29] strategies. Additionally, we evaluate these strategies across three different datasets and three distinct hidden layer sizes (128, 256, and 512) as depicted in Fig. 2. The baseline model for this experiment is established using the centralized training method based on the DeepConvLSTM architecture, with equivalent hidden layer units applied across the datasets.

To ensure consistent result comparability, in line with [7], this project uses an Adam optimizer with a reduced weight decay (1e-6) and learning rate (1e-4), and employ Glorot initialization [30] for network weight setup. During the training, Five-Fold Cross-Validation was applied to calculate loss, enabling the network to adapt to unbalanced datasets. This study primarily investigates differences between centralized and federated training methods and compares outcomes across various federated strategies without excessively adjusting the hyperparameters for the RWHAR, SBHAR, and HHAR datasets. However, since the HHAR dataset samples at 100 Hz, we adjust the size of the convolution filters to 21 to preserve the relationship between filter size and the sliding window duration, ensuring that each filter captures a consistent amount of information across all datasets.

In this project, Hyperledger Fabric [31] will be deployed to structure the decentralized global model consisting of M servers. This decentralized network configuration will include two peer nodes, one order node, and one channel, namely Ledger. The process unfolds as follows: Initially, the global model parameters are proposed to the peer nodes as the inaugural proposal, where they are processed using the Chain Code installed on these nodes. The order node then encapsulates this proposal into the genesis block, which is subsequently dispatched to each peer node for independent verification. Once verified, the proposal is added to the ledger. This procedure is replicated for each round of global model updates, thereby ensuring the traceability, trackability, and security of the model parameters. Additionally, this approach mitigates the risk of single points of failure, enhancing overall system stability.

   \begin{figure}[thpb]
      \centering
      \includegraphics[scale=0.5]{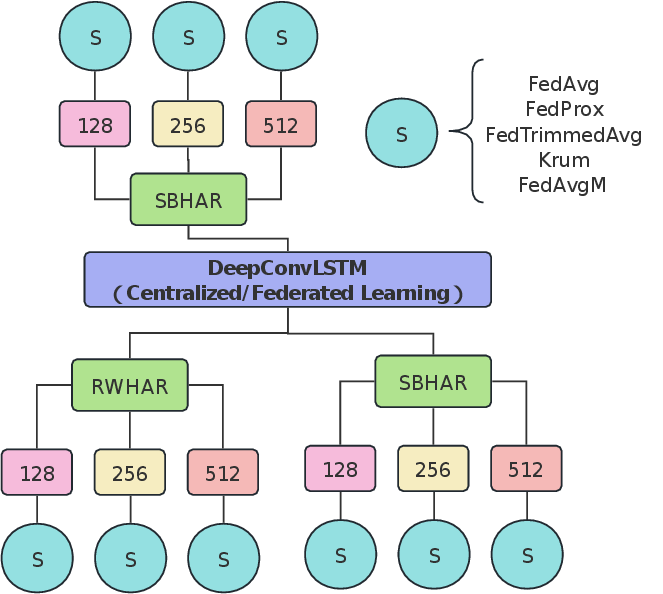} % scale adjust width width=\textwidth
      \caption{CONFIGURATION STRUCTURE OF THE EXPERIMENTAL TASKS}
      \label{figurelabel}
   \end{figure}

\section{RESULTS}

\begin{table*}[ht]
\caption{COMPARISON OF FEDERATED LEARNING WITH CENTRALIZED TRAINING ON HHAR }
\centering
\begin{tblr}{
  colspec={|X[0.6,l]|X[0.6,l]|X[1.1,l]|X[1.1,c]|X[1.1,c]|X[1.1,c]|X[1.1,c]|X[1.1,c]|X[1.1,c]|},
  cells = {c},
  cell{1}{1} = {r=2}{},
  cell{1}{2} = {r=2}{},
  cell{1}{3} = {r=2}{},
  cell{1}{4} = {c=6}{},
  cell{3}{1} = {r=9}{},
  cell{3}{2} = {r=3}{},
  cell{6}{2} = {r=3}{},
  cell{9}{2} = {r=3}{},
  vlines,
  hline{1,3,12} = {-}{},
  hline{2} = {4-9}{},
  hline{4-5,7-8,10-11} = {3-9}{},
  hline{6,9} = {2-9}{},
}
\textbf{Dataset} & \textbf{Hidden Units} & \textbf{Metrics} & \textbf{Centralized Training \& Federated Learning with Five Strategies} &                          &                           &                                 &                        &                           \\
                 &                       &                  & \textit{\textbf{Centralized}}                                            & \textit{\textbf{FedAvg}} & \textit{\textbf{FedProx}} & \textit{\textbf{FedTrimmedAvg}} & \textit{\textbf{Krum}} & \textit{\textbf{FedAvgM}} \\
HHAR             & 128                   & Precision        & 76.24\%                                                                  & 80.61\%                  & \textbf{80.78\%}          & 80.32\%                         & 80.30\%                & \textbf{80.74\%}          \\
                 &                       & Recall           & 76.66\%                                                                  & 80.98\%                  & \textbf{81.10\%}          & 80.75\%                         & 80.64\%                & \textbf{81.06\%}          \\
                 &                       & F1               & 76.33\%                                                                  & 80.70\%                  & \textbf{80.81\%}          & 80.42\%                         & 80.34\%                & \textbf{80.76\%}          \\
                 & 256                   & Precision        & 77.24\%                                                                  & \textbf{81.75\%}         & 81.57\%                   & 80.99\%                         & 81.24\%                & \textbf{81.65\%}          \\
                 &                       & Recall           & 77.55\%                                                                  & \textbf{82.09\%}         & 81.90\%                   & 81.45\%                         & 81.47\%                & \textbf{81.93\%}          \\
                 &                       & F1               & 77.28\%                                                                  & \textbf{81.84\%}         & 81.64\%                   & 81.12\%                         & 81.23\%                & \textbf{81.70\%}          \\
                 & 512                   & Precision        & 78.02\%                                                                  & 82.33\%                  & 82.27\%                   & \textbf{82.34\%}                & 82.29\%                & \textbf{82.37\%}          \\
                 &                       & Recall           & 78.32\%                                                                  & \textbf{82.69\%}         & 82.55\%                   & 82.65\%                         & 82.59\%                & \textbf{82.72\%}          \\
                 &                       & F1               & 78.05\%                                                                  & \textbf{82.43\%}         & 82.34\%                   & 82.42\%                         & 82.34\%                & \textbf{82.46\%}          
\end{tblr}
\end{table*}

In both centralized and federated learning training methods, K-Fold Cross-Validation (specifically Five-Fold) is uniformly applied across each dataset. This approach ensures that every data point serves both as part of the training and the validation set, which is particularly advantageous when data volumes are limited. Additionally, the data preprocessing in federated training is consistent with that of centralized methods across each dataset.

As illustrated in Fig. 2, this study compares the training results of three different hidden layer units (128, 256, and 512) using a single-layer DeepConvLSTM architecture under centralized and federated learning methods (across five different strategies). The results for each strategy represent the average performance of three clients, with outcomes averaged over five runs. This experiment employs precision, recall, and F1-score as the evaluation metric.

Table I displays that all five federated learning strategies outperformed the centralized training approach on HHAR. Notably, FedAvgM achieved superior performance not only compared to centralized training but also excelled over the other strategies under various hidden unit settings. Moreover, with 128 hidden units, FedProx delivered Precision, Recall, and F1 scores of 80.78\%, 81.10\%, and 80.81\% respectively, showing a significant improvement over the other three federated strategies. With 256 hidden units, FedAvg surpassed the others, with increases of 5.84\%, 5.85\%, and 5.90\% in the three metrics respectively. Therefore, when implementing DeepConvLSTM on a single layer using one of these five federated learning strategies, the overall performance on the HHAR dataset exceeded that of the centralized approach, with specific outcomes depending on the number of hidden units.

In the RWHAR training set(Table II), the outcomes of the five federated training methods consistently surpass those of centralized training. Unlike the results seen in the HHAR dataset, for the RWHAR dataset, irrespective of the federated strategy employed, the F1-score for models with 128 hidden layer units consistently outperforms those with 256 and 512 units. For instance, under federation strategy FedAvg, the F1-score improvement at 128 units is 4.49\%, compared to 4.01\% and 4.18\% for 256 and 512 units, respectively. Similarly, in FedAvgM, the F1-score for 128 units rose by 4.56\%, while increases for 256 and 512 units were 4.01\% and 4.24\%, respectively.

\begin{table*}[ht]
\caption{COMPARISON OF FEDERATED LEARNING WITH CENTRALIZED TRAINING ON RWHAR}
\centering
\begin{tblr}{
  colspec={|X[0.6,l]|X[0.6,l]|X[1.1,l]|X[1.1,c]|X[1.1,c]|X[1.1,c]|X[1.1,c]|X[1.1,c]|X[1.1,c]|},
  cells = {c},
  cell{1}{1} = {r=2}{},
  cell{1}{2} = {r=2}{},
  cell{1}{3} = {r=2}{},
  cell{1}{4} = {c=6}{},
  cell{3}{1} = {r=9}{},
  cell{3}{2} = {r=3}{},
  cell{6}{2} = {r=3}{},
  cell{9}{2} = {r=3}{},
  vlines,
  hline{1,3,12} = {-}{},
  hline{2} = {4-9}{},
  hline{4-5,7-8,10-11} = {3-9}{},
  hline{6,9} = {2-9}{},
}
\textbf{Dataset} & \textbf{Hidden Units} & \textbf{Metrics} & \textbf{Centralized Training \& Federated Learning with Five Strategies} &                          &                           &                                 &                        &                           \\
                 &                       &                  & \textbf{\textit{Centralized}}                                            & \textbf{\textit{FedAvg}} & \textbf{\textit{FedProx}} & \textbf{\textit{FedTrimmedAvg}} & \textbf{\textit{Krum}} & \textbf{\textit{FedAvgM}} \\
RWHAR            & 128                   & Precision        & 84.31\%                                                                  & \textbf{88.00\%}         & 87.90\%                   & 87.89\%                         & 87.71\%                & \textbf{87.97\%}          \\
                 &                       & Recall           & 83.71\%                                                                  & 87.55\%                  & \textbf{87.66\%}          & 87.63\%                         & 87.30\%                & \textbf{87.69\%}          \\
                 &                       & F1               & 83.81\%                                                                  & 87.58\%                  & \textbf{87.61\%}          & 87.58\%                         & 87.32\%                & \textbf{87.63\%}          \\
                 & 256                   & Precision        & 85.33\%                                                                  & \textbf{88.51\%}         & 88.46\%                   & 88.41\%                         & 88.25\%                & \textbf{88.51\%}          \\
                 &                       & Recall           & 84.74\%                                                                  & \textbf{88.34\%}         & \textbf{88.29\%}          & 88.22\%                         & 88.15\%                & \textbf{88.29\%}          \\
                 &                       & F1               & 84.85\%                                                                  & \textbf{88.25\%}         & 88.21\%                   & 88.17\%                         & 88.01\%                & \textbf{88.25\%}          \\
                 & 512                   & Precision        & 85.79\%                                                                  & \textbf{89.16\%}         & 89.02\%                   & 89.12\%                         & 89.06\%                & \textbf{89.18\%}          \\
                 &                       & Recall           & 85.43\%                                                                  & 89.08\%                  & 89.08\%                   & \textbf{89.15\%}                & 89.11\%                & \textbf{89.18\%}          \\
                 &                       & F1               & 85.47\%                                                                  & \textbf{89.04\%}         & 88.93\%                   & \textbf{89.04\%}                & 88.98\%                & \textbf{89.09\%}          
\end{tblr}
\end{table*}

\begin{table*}[ht]
\caption{COMPARISON OF FEDERATED LEARNING WITH CENTRALIZED TRAINING ON SBHAR}
\centering
\begin{tblr}{
  colspec={|X[0.6,l]|X[0.6,l]|X[1.1,l]|X[1.1,c]|X[1.1,c]|X[1.1,c]|X[1.1,c]|X[1.1,c]|X[1.1,c]|},
  cells = {c},
  cell{1}{1} = {r=2}{},
  cell{1}{2} = {r=2}{},
  cell{1}{3} = {r=2}{},
  cell{1}{4} = {c=6}{},
  cell{3}{1} = {r=9}{},
  cell{3}{2} = {r=3}{},
  cell{6}{2} = {r=3}{},
  cell{9}{2} = {r=3}{},
  vlines,
  hline{1,3,12} = {-}{},
  hline{2} = {4-9}{},
  hline{4-5,7-8,10-11} = {3-9}{},
  hline{6,9} = {2-9}{},
}
\textbf{Dataset} & \textbf{Hidden Units} & \textbf{Metrics} & \textbf{Centralized Training \& Federated Learning with Five Strategies} &                          &                           &                                 &                        &                           \\
                 &                       &                  & \textbf{\textit{Centralized}}                                            & \textbf{\textit{FedAvg}} & \textbf{\textit{FedProx}} & \textbf{\textit{FedTrimmedAvg}} & \textbf{\textit{Krum}} & \textbf{\textit{FedAvgM}} \\
SBHAR            & 128                   & Precision        & 51.65\%                                                                  & \textbf{54.27\%}         & \textbf{54.20\%}          & 54.16\%                         & 53.39\%                & 53.51\%                   \\
                 &                       & Recall           & 61.59\%                                                                  & 63.55\%                  & \textbf{64.18\%}          & \textbf{64.30\%}                & 63.09\%                & 63.06\%                   \\
                 &                       & F1               & 53.38\%                                                                  & 55.91\%                  & \textbf{55.96\%}          & \textbf{56.02\%}                & 54.93\%                & 55.11\%                   \\
                 & 256                   & Precision        & 53.08\%                                                                  & 55.15\%                  & 55.77\%                   & \textbf{55.88\%}                & 55.29\%                & \textbf{55.90\%}          \\
                 &                       & Recall           & 64.53\%                                                                  & 65.28\%                  & 66.12\%                   & \textbf{66.27\%}                & 65.91\%                & \textbf{66.32\%}          \\
                 &                       & F1               & 54.97\%                                                                  & 57.18\%                  & 57.98\%                   & \textbf{58.09\%}                & 57.57\%                & \textbf{58.12\%}          \\
                 & 512                   & Precision        & 53.70\%                                                                  & \textbf{56.20\%}         & \textbf{55.83\%}          & 55.74\%                         & 55.32\%                & 55.72\%                   \\
                 &                       & Recall           & 65.53\%                                                                  & \textbf{66.76\%}         & 65.89\%                   & 65.92\%                         & 65.51\%                & \textbf{65.99\%}          \\
                 &                       & F1               & 55.87\%                                                                  & \textbf{58.55\%}         & 57.98\%                   & \textbf{57.90\%}                & 57.40\%                & 57.81\%                   
\end{tblr}
\end{table*}

Contrasting with these two datasets, the SBHAR dataset(Table III) does not show a significant improvement in recall. Notably, when employing 512 hidden layer units under Krum, there is even a slight decline of 0.02\%. Specifically, in FedAvg, the recall for 128 units increased by 3.18\%, for 256 by 1.16\%, and for 512 by 1.87\%. In FedProx, the recall enhancements for 128 and 256 hidden units are more pronounced, at 4.2\% and 2.45\% respectively, while the improvement for 512 units is only 0.54\%. Therefore, in the SBHAR dataset, reducing the number of hidden layers significantly enhances performance improvements. This trend is also evident across different federated strategies.

\begin{table*}
\caption{COMPARISON OF AVERAGING THREE DATASETS AND THREE TYPES OF HIDDEN UNITS}
\centering
\begin{tblr}{
  colspec={|X[0.6,l]|X[1.1,c]|X[1.1,c]|X[1.1,c]|X[1.1,c]|X[1.1,c]|X[1.1,c]|},
  cells = {c},
  cell{1}{1} = {r=2}{},
  cell{1}{2} = {c=5}{},
  vlines,
  hline{1,3-6} = {-}{},
  hline{2} = {2-6}{},
}
\textbf{Metrics} & \textbf{Centralized Training \& Federated Learning with Five Strategies} &                  &                        &               &                  \\
                 & \textbf{FedAvg}                                                          & \textbf{FedProx} & \textbf{FedTrimmedAvg} & \textbf{Krum} & \textbf{FedAvgM} \\
Precision        & \textbf{4.75\%}                                                          & 4.74\%           & 4.60\%                 & 4.65\%        & 4.65\%           \\
Recall           & 4.04\%                                                                   & 4.13\%           & \textbf{4.20}\%        & 3.65\%        & 4.02\%           \\
F1               & 4.83\%                                                                   & \textbf{4.85\%}  & 4.25\%                 & 4.25\%        & 4.72\%           
\end{tblr}
\end{table*}

Thus far, we perform a side-by-side comparison of the experimental results, i.e., a comparison of the results of the five federated learning strategies on Precision, Recall, and F1-score based on three different hidden layer units running on three datasets. The results(Table IV) show that FedAvg improves the most significantly on Precision, i.e., 4.75\%. Compared to the other strategies, FedTrimmedAvg performs best on Recall, with an improvement of 0.55 percentage points compared to Krum. And on F1-score, the best strategy for improvement was FedProx, which improved by 4.85\% compared to the centralized training method.

In summary, regardless of the federated learning strategy adopted, results consistently outperform centralized training methods. Specifically, on the HHAR dataset with 512 hidden layer units, the five federated learning strategies yielded average improvements of 5.51\% in precision, 5.52\% in recall, and 5.57\% in F1-score. On the RWHAR dataset with 128 hidden layer units, the strategies averaged improvements of 4.25\% in precision, 4.60\% in recall, and 4.46\% in F1-score. Meanwhile, on the SBHAR dataset with 256 hidden layer units, the improvements were 4.76\% in precision, 2.24\% in recall, and 5.12\% in F1-score. These results highlight nuanced variations across different federated learning strategies and datasets, influenced by the number of hidden layer units employed.

\section{CONCLUSION AND FUTURE WORK}
In conclusion, this paper explores federated transformation and centralized training using the single-layer LSTM architecture proposed by [12], applying the Flower framework to execute five distinct federated learning strategies across three real datasets, each with three varying sizes of hidden units (128, 256, and 512). The experimental results demonstrate that, compared to centralized training methods, the single-layer DeepConvLSTM employing federated learning exhibits improvements in Precision, Recall, and F1-Score to varying extents. Additionally, this project also conducted a horizontal comparison of performance enhancements across different federated strategies, FedAvg, FedTrimmedAvg and FedProx have different levels of improvement in Precision, Recall and F1-score evaluation criteria respectively compared to other strategies.

Furthermore, This paper has replaced the centralized global model in federated learning with a decentralized structure using Hyperledger Fabric. This adjustment not only mitigates the risk of single points of failure but also enhances the tamper-proofing and traceability of model parameters.

In ongoing research, we aim to further bolster security and privacy in the federated learning process. On one hand, we plan to employ advanced encryption protocols, such as homomorphic encryption or secure multi-party computation, during data transmission between clients and the server to protect against theft and reverse engineering of model parameters. On the other hand, we seek to refine the consensus strategy within the server's decentralized structure to enhance the verification efficiency and processing speed of the entire network. Through these improvements, we hope to effectively advance the application of federated learning and blockchain technologies in the wearable sector, significantly enhancing the protection of user privacy and security.

%%%%%%%%%%%%%%%%%%%%%%%%%%%%%%%%%%%%%%%%%%%%%%%%%%%%%%%%%%%%%%%%%%%%%%%%%%%%%%%%


\begin{thebibliography}{99}

\bibitem{c1} G. Saleem, U. I. Bajwa, and R. H. Raza, “Toward human activity recognition: a survey,” Neural Computing and Applications, Oct. 2022, doi: https://doi.org/10.1007/s00521-022-07937-4.
\bibitem{c2} F. Ordóñez and D. Roggen, “Deep Convolutional and LSTM Recurrent Neural Networks for Multimodal Wearable Activity Recognition,” Sensors, vol. 16, no. 1, p. 115, Jan. 2016, doi: https://doi.org/10.3390/s16010115.
\bibitem{c3} D. Roggen et al., “Collecting complex activity datasets in highly rich networked sensor environments,” 2010 Seventh International Conference on Networked Sensing Systems (INSS), Jun. 2010, doi: https://doi.org/10.1109/inss.2010.5573462.
\bibitem{c4} P. Zappi et al., “Activity Recognition from On-Body Sensors: Accuracy-Power Trade-Off by Dynamic Sensor Selection,” Lecture Notes in Computer Science, pp. 17–33, doi: https://doi.org/10.1007/978-3-540-77690-1\_2.
\bibitem{c5} K. Chen, D. Zhang, L. Yao, B. Guo, Z. Yu, and Y. Liu, “Deep Learning for Sensor-based Human Activity Recognition,” ACM Computing Surveys, vol. 54, no. 4, pp. 1–40, May 2021, doi: https://doi.org/10.1145/3447744.
\bibitem{c6} Andrej Karpathy, J. C. Johnson, and L. Fei-Fei, “Visualizing and Understanding Recurrent Networks,” arXiv (Cornell University), Jun. 2015, doi: https://doi.org/10.48550/arxiv.1506.02078.
\bibitem{c7} M. Bock, A. Hoelzemann, M. Moeller, and K. Van Laerhoven, “Improving Deep Learning for HAR with shallow LSTMs,” arXiv.org, Sep. 21, 2021. https://arxiv.org/abs/2108.00702.
\bibitem{c8} Y. Liu et al., “Vertical Federated Learning: Concepts, Advances, and Challenges,” IEEE transactions on knowledge and data engineering, pp. 1–20, Jan. 2024, doi: https://doi.org/10.1109/tkde.2024.3352628.
\bibitem{c9} A. Qammar, A. Karim, H. Ning, and J. Ding, “Securing federated learning with blockchain: a systematic literature review,” Artificial Intelligence Review, Sep. 2022, doi: https://doi.org/10.1007/s10462-022-10271-9.
\bibitem{c10} B. McMahan, E. Moore, D. Ramage, S. Hampson, and B. Aguera y Arcas, "Communication-Efficient Learning of Deep Networks from Decentralized Data," in Proc. 20th Int. Conf. Artif. Intell. Statist. (AISTATS), PMLR 54, 2017, pp. 1273-1282.
\bibitem{c11} J. Konečný, H. Mcmahan, F. Yu, A. Theertha, D. Google, and P. Richtárik, “FEDERATED LEARNING: STRATEGIES FOR IMPROVING COMMUNICATION EFFICIENCY.” Available: https://arxiv.org/pdf/1610.05492.pdf.
\bibitem{c12} T. Li, M. Sanjabi, A. Beirami, and V. Smith, “Fair Resource Allocation in Federated Learning,” arXiv:1905.10497 [cs, stat], Feb. 2020, Available: https://arxiv.org/abs/1905.10497.
\bibitem{c13} K. Bonawitz et al., “Practical Secure Aggregation for Privacy-Preserving Machine Learning,” Proceedings of the 2017 ACM SIGSAC Conference on Computer and Communications Security, Oct. 2017, doi: https://doi.org/10.1145/3133956.3133982.
\bibitem{c14} R. C. Geyer, T. Klein, and M. Nabi, “Differentially Private Federated Learning: A Client Level Perspective,” arXiv:1712.07557 [cs, stat], Mar. 2018, Available: https://arxiv.org/abs/1712.07557.
\bibitem{c15} A. Bulling, U. Blanke, and B. Schiele, “A tutorial on human activity recognition using body-worn inertial sensors,” ACM Computing Surveys, vol. 46, no. 3, pp. 1–33, Jan. 2014, doi: https://doi.org/10.1145/2499621.
\bibitem{c16} G. Gad and Z. Fadlullah, “Federated Learning via Augmented Knowledge Distillation for Heterogenous Deep Human Activity Recognition Systems,” Sensors, vol. 23, no. 1, p. 6, Dec. 2022, doi: https://doi.org/10.3390/s23010006.
\bibitem{c17} Nikola Simić et al., “Enhancing Emotion Recognition through Federated Learning: A Multimodal Approach with Convolutional Neural Networks,” Applied sciences, vol. 14, no. 4, pp. 1325–1325, Feb. 2024, doi: https://doi.org/10.3390/app14041325.
\bibitem{c18} A. Khatoon, “A Blockchain-Based Smart Contract System for Healthcare Management,” Electronics, vol. 9, no. 1, p. 94, Jan. 2020, doi: https://doi.org/10.3390/electronics9010094. 
\bibitem{c19} Q. Liu, Y. Liu, M. Luo, D. He, H. Wang, and K.-K. R. Choo, “The Security of Blockchain-Based Medical Systems: Research Challenges and Opportunities,” IEEE Systems Journal, pp. 1–12, 2022, doi: https://doi.org/10.1109/JSYST.2022.3155156.
\bibitem{c20} D. M. Patel, C. K. Sahu, and R. Rai, “Security in modern manufacturing systems: integrating blockchain in artificial intelligence-assisted manufacturing,” International Journal of Production Research, pp. 1–31, Sep. 2023, doi: https://doi.org/10.1080/00207543.2023.2262050. 
\bibitem{c21} A. Stisen et al., “Smart Devices are Different,” Proceedings of the 13th ACM Conference on Embedded Networked Sensor Systems, Nov. 2015, doi: https://doi.org/10.1145/2809695.2809718.
\bibitem{c22} Timo Sztyler and Heiner Stuckenschmidt, “On-body localization of wearable devices: An investigation of position-aware activity recognition,” Mar. 2016, doi: https://doi.org/10.1109/percom.2016.7456521.
\bibitem{c23} M. Núñez-Regueiro, “Yaşlı Kadınlarda Üreme Sağlığı,” DergiPark (Istanbul University), vol. 1, no. 1, Feb. 2015, doi: https://doi.org/10.1016/j.
\bibitem{c24} D. J. Beutel et al., “Flower: A Friendly Federated Learning Research Framework,” arXiv:2007.14390 [cs, stat], Apr. 2021, Available: https://arxiv.org/abs/2007.14390.
\bibitem{c25} M. H. Brendan, E. Moore, D. Ramage, S. Hampson, and Arcas, Blaise Agüera y, “Communication-Efficient Learning of Deep Networks from Decentralized Data,” arXiv.org, 2016. https://arxiv.org/abs/1602.05629.
\bibitem{c26} T. Li, A. K. Sahu, M. Zaheer, M. Sanjabi, A. Talwalkar, and V. Smith, “Federated Optimization in Heterogeneous Networks,” arXiv:1812.06127 [cs, stat], Apr. 2020, Available: https://arxiv.org/abs/1812.06127.
\bibitem{c27} D. Yin, Y. Chen, K. Ramchandran, and P. Bartlett, “Byzantine-Robust Distributed Learning: Towards Optimal Statistical Rates,” arXiv.org, Feb. 25, 2021. https://arxiv.org/abs/1803.01498.
\bibitem{c28} P. Blanchard, E. M. E. Mhamdi, R. Guerraoui, and J. Stainer, “Byzantine-Tolerant Machine Learning,” arXiv.org, Mar. 08, 2017. https://arxiv.org/abs/1703.02757.
\bibitem{c29} T.-M. H. Hsu, H. Qi, and M. Brown, “Measuring the Effects of Non-Identical Data Distribution for Federated Visual Classification,” arXiv:1909.06335 [cs, stat], Sep. 2019, Available: https://arxiv.org/abs/1909.06335.
\bibitem{c30} X. Glorot and Y. Bengio, “Understanding the difficulty of training deep feedforward neural networks,” proceedings.mlr.press, Mar. 31, 2010. http://proceedings.mlr.press/v9/glorot10a.
\bibitem{c31} E. Androulaki et al., “Hyperledger fabric,” Proceedings of the Thirteenth EuroSys Conference on - EuroSys ’18, 2018, doi: https://doi.org/10.1145/3190508.3190538.
\end{thebibliography}
\end{document}